\documentclass[10pt,conference]{IEEEtran}
\IEEEoverridecommandlockouts
\usepackage{cite}
\usepackage{amsmath,amssymb,amsfonts}
\usepackage{algorithmic}
\usepackage{graphicx}
\usepackage{textcomp}
\usepackage{xcolor}
\def\BibTeX{{\rm B\kern-.05em{\sc i\kern-.025em b}\kern-.08em
    T\kern-.1667em\lower.7ex\hbox{E}\kern-.125emX}}
\begin{document}

\title{Position Paper: Post-Solve Robustness in Decision Engines: Feasible Regions and Smoothness Under Perturbations}

\author{\IEEEauthorblockN{Yi-Xiang Hu}
\IEEEauthorblockA{\textit{University of Science and Technology of China}, Hefei, China,\
 yixianghu@mail.ustc.edu.cn}}

\maketitle

\begin{abstract}
Mixed-Integer Linear Programming (MILP) decision engines routinely output nominally optimal plans for high-stakes industrial systems. Yet deployment rarely matches solve-time assumptions: small perturbations in costs, demands, or resource availability can invalidate feasibility or trigger discontinuous shifts to qualitatively different solutions. We argue that this post-solve robustness gap is a missing layer in today's optimization pipelines and a missing evaluation dimension for learning-enabled decision systems. Rather than replacing robust optimization or stochastic programming, the proposed layer audits a solved incumbent and returns solver-backed evidence about how far that solution can be trusted. We formalize two central objects: (i) an $\epsilon$-near-optimal feasible neighborhood in parameter space, capturing when an incumbent remains feasible and near-optimal under perturbations, and (ii) solution smoothness in decision space, capturing whether nearby alternatives with small combinatorial edits remain competitive. We then synthesize the most relevant partial answers from sensitivity and stability analysis, robust optimization, neighborhood search, adversarial testing, and learning-based enhancements, and articulate an agenda for a unified post-solve robustness layer. Concretely, we call for certified inner approximations around the incumbent, probabilistic robustness estimation with calibrated uncertainty, adversarial robustness margins, and learning-based prediction and explanation aligned with solver-backed verification. We conclude with a compact reporting template and evaluation protocol that would make robustness a first-class output of decision engines.
\end{abstract}


\section{Introduction}
Modern decision support systems, referred to here as \emph{decision engines}, increasingly rely on Mixed-Integer Linear Programming (MILP) solvers to produce plans in logistics, scheduling, finance, and energy. However, deployment rarely matches solve-time assumptions. Forecast errors, execution noise, and model mismatch perturb costs, demands, and resource availability after an incumbent plan is produced. In MILPs, even small perturbations can flip integer assignments, yielding discontinuous changes in discrete decisions that cause feasibility loss or abrupt shifts to qualitatively different solutions.

This deployment gap is familiar to practitioners. A routing plan that is optimal at solve time may become infeasible after a modest capacity reduction; a unit-commitment schedule may remain feasible but require a qualitatively different set of on/off decisions after a small forecast error. In such settings, the nominal objective value is not enough. Users also need to know whether the incumbent remains trustworthy, which perturbations are most dangerous, and whether there are nearby fallback solutions that preserve most of the original value.

We therefore argue that MILP decision engines should not be evaluated or deployed without a \emph{post-solve robustness layer} attached to the solved instance. The layer takes a nominal incumbent $x^*$ and produces a compact robustness report: a certificate or lower bound on feasibility tolerance, a calibrated estimate of failure risk under realistic perturbations, a small set of critical failure modes, and a few high-quality fallback solutions. The intent is not to replace robust optimization (RO), stochastic programming, or receding-horizon control. Those methods act \emph{before} or \emph{during} optimization by redesigning the model or policy. In contrast, the post-solve layer acts \emph{after} optimization by auditing the returned incumbent under a bounded latency budget and presenting the result in a form that supports deployment decisions.

Classical post-optimal sensitivity analysis is well established for linear programs (LPs): one can compute ranges for objective coefficients or right-hand side values within which the optimal basis remains optimal. Extending such analysis to MILPs is difficult because changes in integer variables induce discontinuous shifts in the optimal solution. Early research~\cite{schrage1985sensitivity,dawande2000inference} introduced sensitivity techniques tailored to branch-and-bound, but general MILP sensitivity does not admit LP-style ``optimality ranges.'' Instead, specialized guarantees have been explored. Dawande and Hooker~\cite{dawande2000inference} develop an inference-duality approach that derives linear conditions on parameter changes under which the optimal objective degrades by at most a prescribed tolerance.

A complementary line of work studies stability regions for incumbent MILP solutions. Yi and Lu~\cite{yi2019mixed} analyze how far an optimal MILP solution tolerates parameter variation while remaining valid, and local branching~\cite{fischetti2003local} reveals whether nearby combinatorial alternatives exist. Bridging sensitivity and robustness, the \emph{radius of robust feasibility} captures the largest uncertainty magnitude for which a fixed solution remains feasible; Goberna et al.~\cite{goberna2022radius} survey this notion for linear and integer programs. Taken together, these strands provide valuable partial answers. What remains missing is a unified, solver-attached framework that converts them into a standardized report for deployment.

\begin{figure*}[t]
    \centering
    {%
        \includegraphics[width=0.97\linewidth]{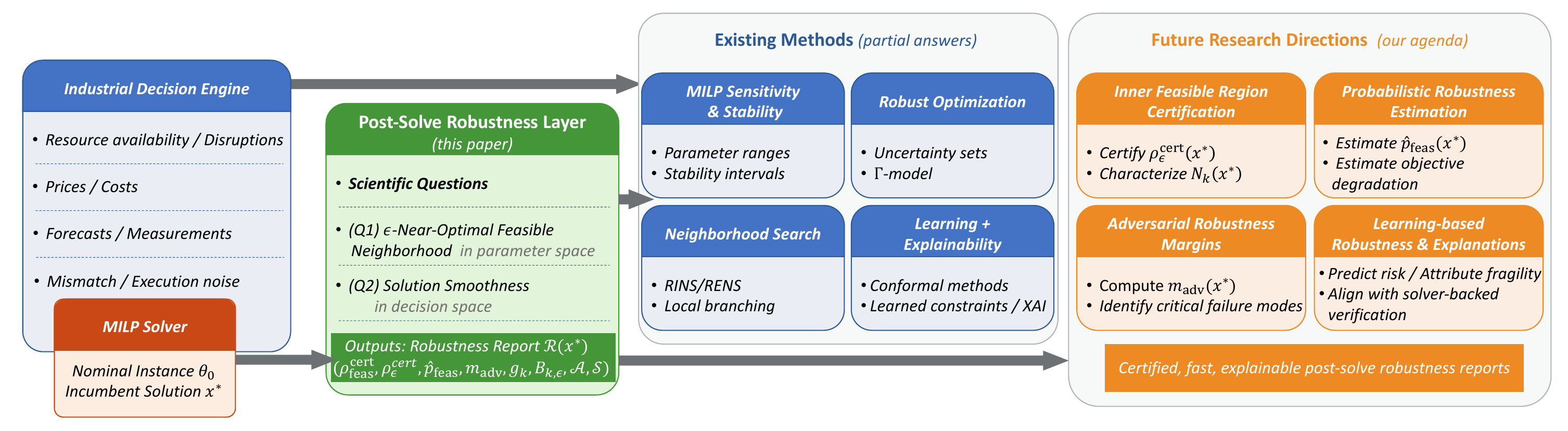}%
    }
    \caption{Existing methods provide partial answers to post-solve robustness, while a dedicated post-solve robustness layer organizes the problem around two scientific questions: the $\epsilon$-near-optimal feasible neighborhood in parameter space and solution smoothness in decision space. The layer returns a structured robustness report $
\mathcal{R}(x^*)$ and motivates four future research directions toward certified, fast, and explainable post-solve robustness reports attached to a solved MILP incumbent.}
    \label{fig:framework}
\end{figure*}

The central claim of this paper is therefore as practical as it is conceptual: robustness should become a first-class solver output rather than an informal afterthought. To make that claim concrete, we formalize two core objects. The first is an $\epsilon$-near-optimal feasible neighborhood in parameter space, which captures when the incumbent remains feasible and near-optimal under perturbations. The second is solution smoothness in decision space, which captures whether nearby alternatives remain competitive under small combinatorial moves. Together, they provide a language for post-solve trustworthiness.

This paper makes four contributions for the position-paper setting. First, it sharpens the boundary between post-solve robustness and adjacent paradigms, especially RO and classical sensitivity analysis. Second, it formalizes the post-solve problem around certified feasible-and-near-optimal neighborhoods and decision-space smoothness. Third, it proposes a concrete reporting template that limits practitioner overload through a tiered summary. Fourth, it outlines a research agenda spanning inner feasible-region certification, probabilistic robustness estimation, adversarial boundary search, and learning-based prediction and explanation. 

The rest of the paper is organized as follows. Section~\ref{sec:problem} defines the problem formally. Section~\ref{sec:background} reviews current approaches and gaps. Section~\ref{sec:future} describes future research directions. Section~\ref{sec:impact} discusses industrial relevance, and Section~\ref{sec:conclusion} concludes.

\section{Problem Definition and Formulation}\label{sec:problem}

We consider a parametric MILP
\begin{equation}
    z(\theta)=\min_{x}\{f(x;\theta): x\in \mathcal{F}(\theta)\cap \mathcal{X}\},
\end{equation}
where $\theta\in\Theta$ collects all perturbable inputs, $\mathcal{F}(\theta)$ denotes the linear constraints induced by $\theta$, and $\mathcal{X}$ encodes integrality restrictions. Let $\theta_0$ be the nominal parameter vector and let $x^*$ be an incumbent optimal solution for $\theta_0$. A \emph{post-solve robustness layer} is a procedure that takes $(\theta_0,x^*)$ and returns a bounded-cost robustness report for deployment.

\vspace{0.5ex}\noindent\textbf{Definition 1 ($\epsilon$-near-optimal feasible neighborhood in parameter space).}
Let $d_\theta$ be a distance on the parameter space. For any radius $\rho\ge 0$, define the perturbation ball around $\theta_0$ by
\begin{equation}
    \mathcal{B}_\rho(\theta_0)=\{\theta\in\Theta: d_\theta(\theta,\theta_0)\le \rho\}.
\end{equation}
To separate pure feasibility from near-optimality, define
\begin{equation}
    \mathcal{U}_{\mathrm{feas}}(x^*)=\{\theta\in\Theta: x^*\in \mathcal{F}(\theta)\cap \mathcal{X}\},
\end{equation}
and, for tolerance $\epsilon\ge 0$,
\begin{equation}
    \mathcal{U}_\epsilon(x^*)=\{\theta\in\Theta: x^*\in \mathcal{F}(\theta)\cap \mathcal{X},\; f(x^*;\theta)\le z(\theta)+\epsilon\}.
\end{equation}
Thus, $\mathcal{U}_{\mathrm{feas}}(x^*)$ is the set of perturbed instances for which the incumbent remains feasible, while $\mathcal{U}_\epsilon(x^*)$ is the set for which the incumbent remains feasible and at most $\epsilon$-suboptimal. The special case $\epsilon=0$ corresponds to exact incumbent optimality under perturbation.

These sets induce two natural certification targets:
\begin{equation}
    \rho^{\mathrm{cert}}_{\mathrm{feas}}(x^*)
    =\sup\{\rho\ge 0: \mathcal{B}_\rho(\theta_0)\subseteq \mathcal{U}_{\mathrm{feas}}(x^*)\},
\end{equation}
and
\begin{equation}
    \rho^{\mathrm{cert}}_{\epsilon}(x^*)
    =\sup\{\rho\ge 0: \mathcal{B}_\rho(\theta_0)\subseteq \mathcal{U}_{\epsilon}(x^*)\}.
\end{equation}
The first quantity is a certified feasibility radius, and the second is a certified near-optimality radius. Both serve as solver-backed lower bounds on how far the incumbent can be trusted under the chosen perturbation model. The feasibility radius is closely related to the robust-feasibility radius surveyed by Goberna et al.~\cite{goberna2022radius}.

\vspace{0.5ex}\noindent\textbf{Definition 2 (Solution smoothness in decision space).}
Let $d_x$ be a distance on feasible solutions; for binary or integer decisions, a Hamming distance is a natural choice. For an integer $k\ge 1$, define the local neighborhood
\begin{equation}
    N_k(x^*)=\{x\in \mathcal{F}(\theta_0)\cap \mathcal{X}: d_x(x,x^*)\le k\}.
\end{equation}
Smoothness should capture whether the incumbent is isolated or supported by nearby competitive alternatives. Two useful quantities are
\begin{equation}
g_k(x^*)=
\begin{cases}
\displaystyle
\min_{x\in N_k(x^*)\setminus\{x^*\}}
& \begin{aligned}[t]
   &\left(f(x;\theta_0)-f(x^*;\theta_0)\right),  \\
   &\text{if }N_k(x^*)\setminus\{x^*\}\neq \emptyset,
  \end{aligned}
\\[1ex]
+\infty, & \text{otherwise}.
\end{cases}
\end{equation}
and
\begin{equation}
    B_{k,\epsilon}(x^*)
    =\left|\left\{x\in N_k(x^*): f(x;\theta_0)\le f(x^*;\theta_0)+\epsilon\right\}\right|.
\end{equation}
Here, $g_k(x^*)$ is the local backup gap and $B_{k,\epsilon}(x^*)$ is the local backup count. A small backup gap and a large backup count indicate a smooth local landscape. A large backup gap or an empty neighborhood indicates fragility. These quantities are more aligned with operational fallback planning than a generic local-optimality score because they measure both the quality and the availability of nearby substitutes.

These definitions suggest a concrete output object. Let $\mathcal D$ denote a stated perturbation distribution over $\Theta$ that reflects the intended deployment regime. Given an incumbent $x^*$, the post-solve layer should return a report
\begin{equation}
    \mathcal{R}(x^*)=
    \big(
    \rho^{\mathrm{cert}}_{\mathrm{feas}},
    \rho^{\mathrm{cert}}_{\epsilon},
    \hat p_{\mathrm{feas}},
    m_{\mathrm{adv}},
    g_k,
    B_{k,\epsilon},
    \mathcal{A},
    \mathcal{S}
    \big),
\end{equation}
where
\begin{equation}
    \hat p_{\mathrm{feas}}(x^*)=
    \Pr_{\theta\sim\mathcal D}\!\big[x^*\in \mathcal{F}(\theta)\cap \mathcal{X}\big]
\end{equation}
is the estimated probability that the incumbent remains feasible under the perturbation model, and
\begin{equation}
    m_{\mathrm{adv}}(x^*)
    =
    \inf\left\{
    d_\theta(\theta,\theta_0):
    \theta\in\Theta,\;
    \theta\notin \mathcal{U}_{\mathrm{feas}}(x^*)
    \right\}
\end{equation}
is the smallest perturbation magnitude that invalidates incumbent feasibility. In the basic report template, $m_{\mathrm{adv}}(x^*)$
refers to the feasibility margin; Section IV-C further distinguishes feasibility and 
$\epsilon$-near-optimality margins. The remaining fields represent a local backup-gap metric, a local backup-count metric, explanatory artifacts $\mathcal{A}$ such as binding constraints or dominant risk factors, and fallback solutions $\mathcal{S}$. Not every deployment requires every field, but this formulation makes the agenda operational: post-solve robustness is not a vague property, but a structured report produced under a post-solve time budget.

The main difficulty is computational. Even membership in $\mathcal{U}_0(x^*)$ can be hard to verify because one must rule out better competing integer solutions under perturbation. Likewise, estimating $g_k(x^*)$ or $B_{k,\epsilon}(x^*)$ may require auxiliary restricted solves. This is precisely why a dedicated research agenda is needed.
\section{Current Approaches and Challenges}\label{sec:background}
Existing methods address fragments of post-solve robustness, yet they rarely compose into a concise, standardized layer attached to a solved MILP. In practice, users need four things at once: trust-region information, failure risk, likely break points, and actionable backups. The current literature usually addresses only one or two of these requirements at a time.

\subsection{MILP Sensitivity Analysis and Stability Regions}
LP sensitivity analysis provides ranges in which the current basis remains optimal. For MILPs, robustness is harder to characterize because small data changes can flip discrete decisions. Early work introduced sensitivity rules tailored to branch-and-bound, tracking how relaxations and node bounds evolve under parameter variations~\cite{schrage1985sensitivity}. Dawande and Hooker proposed inference-based sensitivity analysis that derives linear conditions on perturbations guaranteeing that objective degradation remains within a prescribed tolerance~\cite{dawande2000inference}. These methods are valuable, but they are typically parameter-specific and do not directly produce a deployment-facing report.

Stability-region analysis focuses on the incumbent integer solution itself. Yi and Lu compute coordinate-wise stability intervals by varying one parameter at a time until the incumbent loses optimality~\cite{yi2019mixed}. This yields practical diagnostics, yet multi-parameter coupling remains difficult and discrete switches can occur abruptly. Robust-feasibility-radius analysis shifts attention from optimality to feasibility and measures the largest perturbation magnitude under which a fixed solution remains feasible~\cite{goberna2022radius}. For general MILPs, however, tight radii remain hard to compute at scale.

\subsection{Proactive Robust Optimization Versus Post-Solve Analysis}
RO incorporates uncertainty \emph{during} optimization by enforcing feasibility for all realizations in a specified uncertainty set~\cite{bertsimas2004price,ben2009robust}. This yields built-in protection, and the budgeted uncertainty model of Bertsimas and Sim reduces conservatism through the parameter $\Gamma$~\cite{bertsimas2004price}. RO is indispensable when uncertainty sets are known and the extra computational burden is acceptable.

Our claim is narrower. Post-solve robustness does not replace RO. It complements RO in three common cases: when practitioners still solve a nominal model for speed or modeling simplicity; when uncertainty sets are not available at solve time; and when operators need an audit of a realized incumbent before committing to execution. Under this view, RO changes the optimization problem, whereas post-solve robustness evaluates the returned solution under an explicit latency budget and communicates the result in decision-oriented terms.

\subsection{Neighborhood Search and Inner Feasible Approximations}
Neighborhood search probes local structure around the incumbent. Local branching restricts Hamming distance to the incumbent and re-solves the MILP to find nearby alternatives or certify local optimality within that neighborhood~\cite{fischetti2003local}. Related strategies such as RINS and RENS construct neighborhoods by fixing subsets of variables and re-optimizing the remainder~\cite{danna2005exploring,berthold2007rens}. These methods are powerful for discovering nearby alternatives, but they are usually used as improvement heuristics rather than as standardized robustness diagnostics.

More formal work studies inner approximations that guarantee feasibility around an integer point~\cite{fischetti2010heuristics}. These constructions align closely with our notion of solver-backed certification. The gap is not conceptual possibility, but packaging: current tools are not usually exposed as a compact certificate, a risk score, and a ranked fallback list suitable for a human operator.

\subsection{Simulation and Worst-Case Evaluation}
Scenario-based evaluation tests the incumbent under sampled perturbations to estimate feasibility rates and objective variation. This is practical and distribution-aware, but it provides limited guarantees. Worst-case robustness evaluation searches for an adversarial perturbation within prescribed bounds and connects naturally to robust-optimization separation~\cite{bertsimas2012adaptive}. It can identify concrete breaking scenarios, but the resulting auxiliary problems may themselves be expensive.

\subsection{Learning-Based Enhancements and Explainability}
Learning-based approaches can accelerate robustness assessment, fill in missing structure, and improve interpretability. Constraint-learning surveys formalize how data augments optimization models while controlling feasibility risk~\cite{fajemisin2024optimization}. Conformal mixed-integer constraint learning provides probabilistic feasibility guarantees for learned constraints~\cite{ovalle2025conformal}. Related work integrates neural constraints and decomposition ideas for scalable mixed-integer optimization~\cite{zeng2026scalable,Hu_Wu_Li_Zhao_Li_2025}. On the interaction side, recent systems explore explaining optimization outcomes to practitioners and generating contrastive explanations via constraint reasoning~\cite{chen2025optichat,lera2025exploiting}, while surveys discuss explainable AI in broader decision-making pipelines~\cite{danach2025toward}. These directions improve usability, but they must remain aligned with solver-backed verification to avoid unsupported robustness claims.

\subsection{Why Existing Pieces Do Not Yet Form a Solver Output}
The missing piece is integration. Sensitivity analysis offers local structure but struggles with joint perturbations. RO offers protection but requires uncertainty sets and changes the optimization problem. Neighborhood search finds backups but does not standardize robustness metrics. Simulation and adversarial testing expose failures but may lack certificates or latency discipline. Learning-based methods accelerate assessment and explanation, but they require calibration and verification. A post-solve robustness layer must combine these pieces into a budgeted workflow and expose a small number of operator-facing outputs.

\subsection{A Call to Action: Robustness as a First-Class Output and Evaluation Protocol}\label{sec:robustness_output}
We call for a shift from treating MILP solvers as black boxes that return a single incumbent to treating them as decision engines that return \emph{an incumbent plus a robustness report}. To avoid overwhelming practitioners, the report should be tiered.

\textbf{Tier 1: compact summary.} Four scalar or short-form outputs should appear by default: (i) a certified lower bound on feasibility tolerance, (ii) a calibrated risk estimate under a stated perturbation model, (iii) a smallest adversarial perturbation or ranked critical failure mode, and (iv) one to three near-optimal fallback solutions. Concretely, Tier 1 may expose $(\rho^{\mathrm{cert}}_{\mathrm{feas}},
    \hat p_{\mathrm{feas}},
    m_{\mathrm{adv}},
    \mathcal{S}_{\mathrm{top}})$, where 
$\mathcal{S}_{\mathrm{top}}\subseteq \mathcal{S}$ contains one to three fallback solutions.

\textbf{Tier 2: diagnostic details.} For analysts, the engine should expose binding constraints, dominant perturbation directions, neighborhood-search statistics, and sensitivity traces.

\textbf{Tier 3: full audit.} For debugging or offline model refinement, the engine may return deeper adversarial traces, scenario rollouts, or solver logs.

This structure also addresses standardization. We advocate a \emph{core metric layer} shared across domains and a \emph{domain policy layer} defined by application owners. The core layer contains solver-agnostic quantities such as feasibility radius, adversarial margin, local backup gap, and empirical failure probability. The domain layer specifies what counts as acceptable risk, which perturbation norms matter, and what action should follow a brittle report. This split allows logistics, energy, and finance to compare methods on common robustness axes while preserving domain-specific safety thresholds.

Finally, every report should be produced under an explicit post-solve time budget $\tau_{\mathrm{post}}$. Some fields may be certified exactly, some may be lower bounds, and some may be statistical estimates. This is a feature rather than a defect: a budgeted, anytime report is more useful in practice than an ideal report that arrives too late.
\subsection{Evaluation Protocol}
A post-solve robustness layer should be evaluated as an attached decision service rather than as a standalone predictor. Starting from nominal MILP instances and incumbent solutions $x^*$, one generates perturbed instances from a stated deployment model $D$ over $\Theta$. For each nominal solve, the post-solve layer returns a report $R(x^*)$ under a post-solve budget $\tau_{\mathrm{post}}$, and this report is compared against the realized outcomes on the perturbed instances.

Four dimensions are central. First, \emph{effectiveness} measures feasibility retention, near-optimality retention, and fallback quality under perturbation. Second, \emph{calibration} compares estimated risks, such as $\hat p_{\mathrm{feas}}(x^*)$, with empirical frequencies and checks whether certified radii such as $\rho_{\mathrm{feas}}^{\mathrm{cert}}$ and $\rho_{\epsilon}^{\mathrm{cert}}$ match observed coverage. Third, \emph{efficiency} measures latency, auxiliary solver effort, and the trade-off between report completeness and time budget. Fourth, \emph{actionability} measures whether the report improves deployment decisions relative to returning the nominal incumbent alone.

\section{Future Research Directions}\label{sec:future}
We outline four promising directions for turning the above agenda into a practical post-solve layer.

\subsection{Inner Approximation of Feasible Regions Around Solutions}
A first direction is to characterize the local feasible region around the incumbent $x^*$. 
The goal is to construct an inner approximation of the feasible set near 
$x^*$
 and to quantify, through neighborhoods such as 
$N_k(x^*)$, whether the incumbent is supported by nearby alternatives with comparable objective values. This turns local structure into an explicit robustness object rather than an implicit by-product of solving.

A practical mechanism is to impose neighborhood constraints and re-solve, as in local branching~\cite{fischetti2003local}. Increasing the neighborhood size $k$ reveals how objective quality changes as one moves away from $x^*$ and whether the feasible set around $x^*$ is dense or sparse. Efficient exploration can build on RINS and RENS~\cite{danna2005exploring,berthold2007rens} to discover alternatives of similar quality. A complementary track is to derive certified inner regions through polyhedral constructions~\cite{fischetti2010heuristics}. An important open problem is balancing certificate strength, interpretability, and computational cost so that local robustness reporting becomes routine for large MILPs.

\subsection{Probabilistic Robustness Estimation}
Worst-case guarantees are valuable, but many deployments accept solutions that succeed with high probability under realistic uncertainty. A second direction is therefore to estimate quantities such as $\hat p_{\mathrm{feas}}(x^*)$ and the distribution of objective degradation under a calibrated perturbation model. Scenario evaluation provides a baseline, but future work should improve sample efficiency, concentrate computation near failure boundaries, and provide principled uncertainty quantification.

This direction is especially relevant when exact certification is too expensive. A well-calibrated probability of success is often operationally meaningful, provided the perturbation model is explicit and the estimate is accompanied by confidence information. It also creates a natural interface with online monitoring and recurring solves.
\subsection{Adversarial Perturbation and Robustness Margins}

Complementing probabilistic evaluation, an adversarial perspective seeks the smallest perturbation that breaks the incumbent. Using the notation of Section~\ref{sec:problem}, the natural target is the smallest parameter-space displacement that moves the instance outside the trusted region of $x^*$. For a tolerance $\epsilon\ge 0$, define the adversarial near-optimality margin
\begin{equation}
    m_{\mathrm{adv}}^{\epsilon}(x^*)
    =
    \inf\left\{
    d_\theta(\theta,\theta_0):
    \theta\in\Theta,\;
    \theta\notin \mathcal{U}_{\epsilon}(x^*)
    \right\}.
\end{equation}
Equivalently, $m_{\mathrm{adv}}^{\epsilon}(x^*)$ is the smallest perturbation magnitude under which the incumbent becomes infeasible or more than $\epsilon$-suboptimal. When only feasibility matters, one obtains the feasibility margin
\begin{equation}
    m_{\mathrm{adv}}^{\mathrm{feas}}(x^*)
    =
    \inf\left\{
    d_\theta(\theta,\theta_0):
    \theta\in\Theta,\;
    \theta\notin \mathcal{U}_{\mathrm{feas}}(x^*)
    \right\}.
\end{equation}
These quantities provide interpretable robustness margins. They also identify adversarial instances
\begin{equation}
    \theta_{\mathrm{adv}}^{\epsilon}\in
    \arg\min_{\theta\in\Theta,\;\theta\notin \mathcal{U}_{\epsilon}(x^*)}
    d_\theta(\theta,\theta_0),
\end{equation}
or feasibility-only analogues, which expose the most vulnerable direction in parameter space.

This direction is related to parametric optimization~\cite{yi2019mixed}, but it targets coupled multi-parameter failures rather than one-parameter thresholds. Feasibility margins can begin with slack-based diagnostics and solver-assisted verification. Near-optimality margins are harder because one must identify perturbations under which competing solutions overtake $x^*$. Practical approaches therefore combine restricted adversarial search, near-optimal alternative enumeration, and targeted auxiliary solves. The output should remain concise: an estimated breaking margin, a representative adversarial instance, the associated dominant constraints or risk factors, and a recommended response such as repair, re-optimization, or operator escalation.

\subsection{Learning-Based Robustness Prediction and Explanation}
A final direction is to integrate machine learning so that post-solve evaluation and communication become faster and more actionable. Surrogate models can predict robustness indicators from instance features, incumbent features, and solver artifacts such as slack profiles or relaxation information. This supports rapid screening when similar MILPs are solved repeatedly.

Learning can also reduce model mismatch by augmenting constraints or penalties from data. Conformal methods provide one principled route for incorporating learned components with statistical guarantees~\cite{ovalle2025conformal}, and broader work on optimization with learned components frames this interface cleanly~\cite{fajemisin2024optimization}. The crucial requirement is that learned estimates remain coupled to solver-backed verification.

Finally, learning should improve explanations. Robustness reports are most useful when they say not only that a solution is brittle, but why it is brittle and what can be done next. Contrastive explanation mechanisms based on infeasibility reasoning provide a promising foundation~\cite{lera2025exploiting}. In practice, this means prioritizing the constraints, parameters, and structural choices most responsible for fragility, while still grounding the final claim in explicit solver checks.

\section{Industrial Impact and Use Cases}\label{sec:impact}
Post-solve robustness is driven by deployment requirements. Industrial decision engines operate under forecast error, execution noise, and model mismatch. In these settings, an operator does not only ask whether a solution is optimal; the operator asks whether it is safe to execute, how likely it is to fail, and what to do if it is brittle. Across domains, the common bottleneck is not only solving the nominal MILP, but deciding whether the solved incumbent is safe to execute under bounded uncertainty.

\subsection{Supply Chain and Logistics}
Logistics applications such as routing, network design, and inventory planning face uncertainty in demand, travel time, and cost. RO can produce plans that remain feasible under demand fluctuations~\cite{hu2023two}. In pipeline transportation, Moradi and MirHassani show that $\Gamma$-robust scheduling improves feasibility under demand uncertainty~\cite{moradi2016robust}. A post-solve layer complements these methods by quantifying how much a nominal plan can absorb before repair becomes necessary and by surfacing limited-change fallback plans.

\subsection{Energy Systems}
Power systems require schedules that tolerate uncertainty in load and renewable generation. Robust formulations for unit commitment improve feasibility under forecast error and reduce reliance on emergency actions~\cite{yang2021robust,bertsimas2012adaptive}. A post-solve robustness layer adds a deployment-facing audit: how far the current commitment can drift before violating reserve or ramping constraints, which forecast components are most dangerous, and whether low-regret recourse plans exist nearby.

\subsection{Manufacturing and Scheduling}
Manufacturing and project scheduling face variability in processing times and resource availability\cite{hu2026ischeduler}. Robust scheduling introduces slack and flexibility to avoid infeasibility cascades under disruptions~\cite{yang2021robust}. In practice, however, managers often prefer minimal schedule edits over complete re-optimization. This makes decision-space smoothness especially relevant because it quantifies whether small repairs preserve most of the nominal objective.

\subsection{Finance and Portfolio Optimization}
Portfolio selection via MILP or MIQP is sensitive to uncertainty in returns and market conditions. Robust and learning-augmented formulations aim to preserve feasibility and reduce instability under adverse outcomes~\cite{fajemisin2024optimization,ovalle2025conformal}. A post-solve layer complements these formulations by identifying how quickly the incumbent loses feasibility or near-optimality as assumptions drift and by explaining which exposures drive fragility.

\section{Conclusion}\label{sec:conclusion}
This paper argues that nominal optimality is an incomplete output for high-stakes MILP decision engines. We sharpened the notion of post-solve robustness as a solver-attached audit layer that complements rather than replaces robust optimization and classical sensitivity analysis. We formalized two core objects, an $\epsilon$-near-optimal feasible neighborhood in parameter space and solution smoothness in decision space, and translated them into a concrete robustness-report template built around certificates, risk estimates, adversarial margins, and fallback solutions.

The main message is actionable. When a solver reports that an incumbent is brittle, the system should not stop at that warning. It should also indicate the dominant failure mode, expose nearby repairs when they exist, and signal when escalation to re-optimization or a more explicitly robust model is necessary. This is the practical meaning of making robustness a first-class output.

We therefore view post-solve robustness as both a research problem and an evaluation problem. Scientifically, it requires new methods for certified local analysis, probabilistic estimation, adversarial search, and learning-based explanation. Operationally, it requires budgeted reporting standards that are informative without overwhelming practitioners. At a minimum, future decision engines should expose a certified tolerance, a calibrated risk estimate, a breaking scenario, and a limited-change fallback recommendation alongside the incumbent. Establishing this interface would improve the reliability, explainability, and deployability of optimization-based decision engines.
\section*{Acknowledgments}

 I acknowledge Shang-Hua Teng, Xiang-Yang Li, Feng Wu, and the anonymous reviewers for their comments and helpful feedback.

\bibliographystyle{IEEEtran}
\bibliography{refer}

\end{document}